\documentclass{article}
\usepackage{ijcai23}

\usepackage{times}
\usepackage{soul}
\usepackage{url}
\usepackage[hidelinks]{hyperref}
\usepackage[utf8]{inputenc}
\usepackage[small]{caption}
\usepackage{graphicx}
\usepackage{amsmath}
\usepackage{amsthm}
\usepackage{booktabs}
\usepackage{algorithm}
\usepackage{algorithmic}
\usepackage{algorithmic}
\usepackage{booktabs}
\usepackage{graphicx}
\usepackage{bbding}
\usepackage{pifont}
\usepackage{subfigure}
\usepackage{amsmath}
\usepackage{float}
\usepackage{xcolor}         % colors
\title{Video Object Segmentation in Panoptic Wild Scenes}

\author{
Yuanyou Xu$^{1,2}$\footnotemark[2]
\and
Zongxin Yang$^{1}$\and
Yi Yang$^{1}$\footnotemark[3]
\affiliations
$^1$ReLER, CCAI, Zhejiang University\\
$^2$Baidu Research\\
\emails
\{yoxu, zongxinyang, yangyics\}@zju.edu.cn
}
\begin{document}
\maketitle
% % \pagestyle{plain}
% % \usepackage{nopageno}
% \twocolumn[{
% \renewcommand\twocolumn[1][]{#1}
% \maketitle
% \begin{center}
% \setlength{\abovecaptionskip}{8pt} 
% \setlength{\belowcaptionskip}{0pt} 
% \centering
% % \includegraphics[height=2.4in, width=6.5in]{cvprToeccv/big_first.png}
% \includegraphics[width=1\textwidth]{pics/main-4.pdf}
% \captionof{figure}{
% The figures illustrate challenges of video object segmentation in panoptic wild scenes in VIPOSeg dataset. 
% In crowded scenes, motion and occlusion are sometimes extremely complex. 
% In addition, numerous objects are challenging to the efficiency of VOS models. 
% Objects on various scales are also difficult to deal with, especially small objects. 
% As for object classes, VIPOSeg contains seen/unseen classes and thing/stuff classes. 
% VOS models need to not only generalize from seen to unseen classes, but also learn to process both thing and stuff.}
% \label{fig1}
% \end{center}}]

\begin{figure*}[t]
\centering
\includegraphics[width=0.98\textwidth]{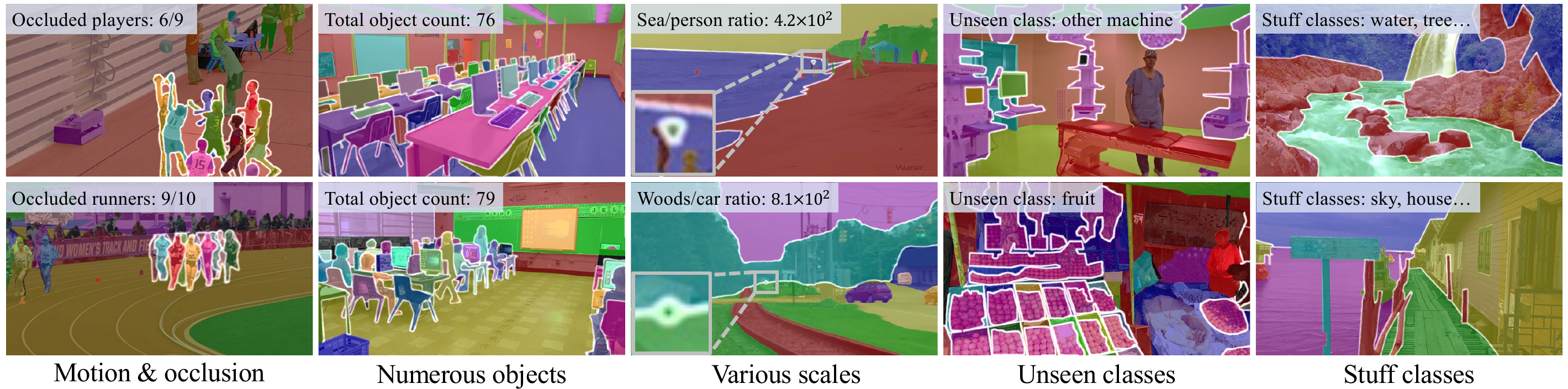}
\caption{The figures illustrate challenges of video object segmentation in panoptic wild scenes in VIPOSeg dataset. 
In crowded scenes, motion and occlusion are sometimes extremely complex. 
In addition, numerous objects are challenging to the efficiency of VOS models. 
Objects on various scales are also difficult to deal with, especially small objects. 
As for object classes, VIPOSeg contains seen/unseen classes and thing/stuff classes. 
VOS models need to not only generalize from seen to unseen classes, but also learn to process both thing and stuff.}
\label{fig1}
\end{figure*}

\begin{abstract}
\renewcommand{\thefootnote}{\fnsymbol{footnote}}
\footnotetext[2]{Yuanyou Xu worked on this at his Baidu Research internship.}
\footnotetext[3]{Yi Yang is the corresponding author.}
\renewcommand*{\thefootnote}{\arabic{footnote}}
In this paper, we introduce semi-supervised video object segmentation (VOS) to panoptic wild scenes and present a large-scale benchmark as well as a baseline method for it. Previous benchmarks for VOS with sparse annotations are not sufficient to train or evaluate a model that needs to process all possible objects in real-world scenarios. Our new benchmark (VIPOSeg) contains exhaustive object annotations and covers various real-world object categories which are carefully divided into subsets of thing/stuff and seen/unseen classes for comprehensive evaluation. Considering the challenges in panoptic VOS, we propose a strong baseline method named panoptic object association with transformers (PAOT), which associates multiple objects by panoptic identification in a pyramid architecture on multiple scales. Experimental results show that VIPOSeg can not only boost the performance of VOS models by panoptic training but also evaluate them comprehensively in panoptic scenes. Previous methods for classic VOS still need to improve in performance and efficiency when dealing with panoptic scenes, while our PAOT achieves SOTA performance with good efficiency on VIPOSeg and previous VOS benchmarks. PAOT also ranks 1$^{st}$ in the VOT2022 challenge. Our dataset and code are available at \url{https://github.com/yoxu515/VIPOSeg-Benchmark}.
\end{abstract}

\section{Introduction}

Video object segmentation (VOS) is a fundamental task in computer vision. In this paper, we focus on semi-supervised video object segmentation, which aims to segment all target objects specified by reference masks in video frames. Although VOS has been well studied in recent years, there are still limitations in previous benchmark datasets. Firstly, previous VOS datasets only provide limited annotations. The annotations of commonly used datasets for VOS, YouTube-VOS \cite{xu2018youtube} and DAVIS \cite{pont20172017} are spatially sparse, with a few objects annotated for most video sequences. Secondly, the classes of YouTube-VOS only include countable thing objects. While in the real world, many scenes may contain dozens of objects and other stuff classes like `water' and `ground'. Obviously these datasets can't cover such scenarios. As a consequence, previous datasets are not able to train VOS models thoroughly and evaluate models comprehensively.

To this end, we study VOS in panoptic scenes as panoptic VOS and present a dataset named VIdeo Panoptic Object Segmentation (VIPOSeg). VIPOSeg is built on VIPSeg \cite{miao2022large}, a dataset for video panoptic segmentation. We re-split the training and validation set and convert the panoptic annotations in VIPSeg to VOS format. Beyond classic VOS, we make thing/stuff annotations for objects available as a new panoptic setting. VIPOSeg dataset is qualified to play the role of a benchmark for panoptic VOS. First, VIPOSeg provides annotations for all objects in scenes. Second, a variety of object categories are included in VIPOSeg. The large diversity of classes and density of objects help to train a model with high robustness and generalization ability for complex real-world applications. For model evaluation, we divide object classes into thing/stuff and seen/unseen subsets. A model can be comprehensively evaluated on these class subsets. In addition, VIPOSeg can also evaluate the performance decay of a model as the number of objects increases.

Challenges also emerge when a model tries dealing with panoptic scenes (Figure \ref{fig1}). The large number of objects causes occlusion and efficiency problem, and various scales and diversity of classes require high robustness. In order to tackle the challenges, we propose a strong baseline method for panoptic object association with transformers (PAOT), which uses decoupled identity banks to generate panoptic identification embeddings for thing and stuff, and uses a pyramid architecture with efficient transformer blocks to perform multi-scale object matching. PAOT achieves superior performance with good efficiency and ranks 1$^{st}$ in both short-term/real-time tracking and segmentation tracks in the VOT2022 challenge \cite{kristan2023tenth}.

% Secondly, most previous VOS methods \cite{oh2019video,yang2020collaborative,cheng2021rethinking} process objects separately in multi-object scenario. The mask for each target object is predicted and then all masks are merged together in a post-processing manner. Such a way can result in inefficiency when the number of objects becomes large, and the association relationship among objects can't be fully captured. Therefore, previous methods may encounter performance and efficiency problems in complex real-world multi-object scenarios.

In summary, our contributions are three-fold:
\begin{itemize}
    \item We introduce panoptic VOS, and present a new benchmark VIPOSeg, which provides exhaustive annotations and includes seen/unseen and thing/stuff classes.
    % Besides, classic/panoptic settings and various metrics are designed for comprehensive evaluation.
    \item Considering the challenges in panoptic VOS, we propose a strong baseline PAOT, which consists of the decoupled identity banks for thing and stuff, and a pyramid architecture with efficient long-short term transformers.
    \item Experimental results show that VIPOSeg is more challenging than previous VOS benchmarks, while our PAOT models show superior performance on the new VIPOSeg benchmark as well as previous benchmarks.
\end{itemize}

\section{Related Work}
\paragraph{Semi-supervised video object segmentation.}
% As an early branch, online VOS methods \cite{caelles2017one,voigtlaender2017online,maninis2018video,yang2018efficient,meinhardt2020make} fine-tune a segmentation model on given masks for each video. Another promising branch is matching-based VOS methods \cite{shin2017pixel,hu2018videomatch,oh2018fast,wang2019ranet,voigtlaender2019feelvos,yang2020collaborative}, which constructs the embedding space to measure the distance between a pixel and the given object. STM \cite{oh2019video} introduces the memory networks to video object segmentation and models the matching as space-time memory reading. Later works \cite{seong2020kernelized,liang2020video,hu2021learning,cheng2021rethinking} improve STM by better memory reading or matching methods. A multi-object identification mechanism is proposed in AOT \cite{yang2021associating,yang2021towards} to process all the target objects simultaneously. We build our method based on AOT and propose solutions to tackle challenges in panoptic scenes.
As an early branch, online VOS methods \cite{caelles2017one,yang2018efficient,meinhardt2020make} fine-tune a segmentation model on given masks for each video. Another promising branch is matching-based VOS methods \cite{shin2017pixel,voigtlaender2019feelvos,yang2020collaborative}, which constructs the embedding space to measure the distance between a pixel and the given object. STM \cite{oh2019video} introduces the memory networks to video object segmentation and models the matching as space-time memory reading. Later works \cite{seong2020kernelized,cheng2021rethinking} improve STM by better memory reading strategies. A multi-object identification mechanism is proposed in AOT \cite{yang2021associating,yang2021towards,yang2022decoupling} to process all the target objects simultaneously. This strategy is adopted in our framework to model the relationship between multiple objects, and we further propose solutions for other challenges in panoptic scenes.

\paragraph{Multi-scale architectures for VOS.}
CFBI+ \cite{yang2021collaborative} proposes a multi-scale foreground and background integration structure, and a hierarchical multi-scale architecture is proposed in HMMN \cite{seong2021hierarchical}. In this work, we also propose a multi-scale architecture, while the matching is performed sequentially in our pyramid architecture but not individually (CFBI+) or with guidance (HMMN). The design of our method is inspired by general transformer backbones \cite{wang2021pyramid,liu2021swin} but ours is for feature matching across multiple frames on both spatial and temporal dimensions but not feature extraction on static images.

\begin{table}[]
% \scriptsize
\fontsize{8pt}{9pt} \selectfont
\setlength{\tabcolsep}{1pt}
\renewcommand{\arraystretch}{1.2}
\resizebox{\columnwidth}{!}{%
\begin{tabular}{|c|c|c|c|c|c|c|c|}
\hline
Dataset                          & Task    & Video         & $T$/s           & Class      & Unseen    & Stuff     & Obj./Video     \\ \hline
DAVIS \cite{pont20172017}        & VOS     & 150           & 2.9           & -            & \ding{55} & \ding{55} & 2.51           \\ \hline
YouTube-VOS \cite{xu2018youtube} & VOS     & \textbf{4453} & 4.5           & 94           & \ding{51} & \ding{55} & 1.64           \\ \hline
UVO \cite{wang2021unidentified}  & OWOS    & 1200          & 3.0             & open         & \ding{51} & \ding{55} & 12.29          \\ \hline
OVIS \cite{qi2022occluded}       & VOS/VIS & 901           & \textbf{12.8} & 25           & \ding{55} & \ding{55} & 5.80           \\ \hline
VIPSeg \cite{miao2022large}      & VPS     & 3536          & 4.8           & 124          & \ding{55} & \ding{51} & 13.26$^\ast$    \\ \hline
\textbf{VIPOSeg}                 & VOS     & 3149          & 4.3           & \textbf{125} & \ding{51} & \ding{51} & \textbf{13.26} \\ \hline
\end{tabular}%
}
\caption{Detailed comparison of related datasets. Obj./Video stands for the average object number per video. $T$ is the average video duration time. $^\ast$For VIPSeg, test set is not included when calculating average object number because it is not public.}
\label{tab:datasets}
\end{table}

\paragraph{Video panoptic segmentation.}
Among the tasks for video segmentation \cite{zhou2022survey,li2023transformer}, video panoptic segmentation (VPS) \cite{kim2020video} is also related to our panoptic VOS. VPS methods \cite{woo2021learning,li2022video,kim2022tubeformer} manage to predict object classes and instances for all pixels in each frame of a video, while in panoptic VOS all objects are defined by reference masks when they first appear. Although they both consider thing and stuff, panoptic VOS is class agnostic and can generalize to arbitrary classes. In addition, most VPS datasets like Cityscapes-VPS \cite{cordts2016cityscapes} and KITTI-STEP \cite{weber2021step} only cover street scenes with limited object categories. VIPSeg \cite{miao2022large} is the first large-scale VPS dataset in the wild.

\paragraph{Related datasets.}
Detailed comparison of related datasets can be found in Table \ref{tab:datasets}, which also covers some datasets beyond VOS. DAVIS \cite{pont20172017} is a small VOS dataset containing 150 videos with sparse object annotations. YouTube-VOS \cite{xu2018youtube} is a large-scale VOS dataset containing 4453 video clips and 94 object categories. OVIS dataset \cite{qi2022occluded} focuses on heavy occlusion problems in video segmentation, in which the 901 video clips mainly include multiple occluded instances. 
UVO \cite{wang2021unidentified} is for open world object segmentation and has much denser annotations than YouTube-VOS. VIPSeg \cite{miao2022large} is a large-scale dataset for video panoptic segmentation in the wild. We build our VIPOSeg dataset based on VIPSeg and details are in the following section.

% \subsubsection{VIS/VPS Datasets}
% Some datasets for relevant tasks, like video instance segmentation and video panoptic segmentation, may also help to train or evaluate VOS models. YouTube-VIS \cite{yang2019video} is a large-scale benchmark dataset for video instance segmentation tasks, which shares a part of videos and annotations with YouTube-VOS. OVIS dataset \cite{qi2022occluded} is proposed to tackle heavy cocclusion problems in VIS. Each video clip in it mainly includes multiple occluded instances. UVO \cite{wang2021unidentified} is for open world object segmentation and has much denser annotations than YouTube-VIS or YouTube-VOS. The open world object segmentation (OWOS) task is class-agnostic so there is no specific class taxonomy in UVO. VIPSeg \cite{miao2022large} is a large-scale dataset for video panoptic segmentation in the wild. build our VIPOSeg dataset based on VIPSeg and details are in the following section.

\section{Benchmark}

% \begin{table}[]
% \centering
% \scriptsize
% \setlength{\tabcolsep}{2pt}
% \renewcommand{\arraystretch}{0.8}
% \resizebox{\columnwidth}{!}{%
% \begin{tabular}{|c|c|c|c|c|c|c|}
% \hline
% Dataset     & Task    & Videos & Classes & Unseen       & Stuff        & Obj/Video \\ \hline
% DAVIS \cite{pont20172017}      & VOS     & 150    & -       & \ding{55} & \ding{55} & 2.51    \\ \hline
% YouTube-VOS \cite{xu2018youtube} & VOS     & \textbf{4453}   & 94      & \ding{51}   & \ding{55} & 1.64    \\ \hline
% UVO \cite{wang2021unidentified}        & OWOS    & 1200   & open    & \ding{51}   & \ding{55} & 12.29   \\ \hline
% OVIS \cite{qi2022occluded}       & VOS/VIS & 901    & 25      & \ding{55} & \ding{55} & 5.80     \\ \hline
% VIPSeg \cite{miao2022large}     & VPS     & 3536   & 124     & \ding{55} & \ding{51}   & 13.26$^\ast$   \\ \hline
% \textbf{VIPOSeg}     & VOS    & 3149   & \textbf{124}     & \ding{51}   & \ding{51}   & \textbf{13.26}   \\ \hline
% \end{tabular}%
% }
% \caption{Details of datasets for VOS. Obj/Video stands for the average object number per video. $\ast$ For VIPSeg, test set is not included when calculating average object number because it is not public.}
% \label{tab:datasets}
% \end{table}

% \usepackage{graphicx}

\begin{figure}[t]
% \centering
\includegraphics[width=1.0\columnwidth]{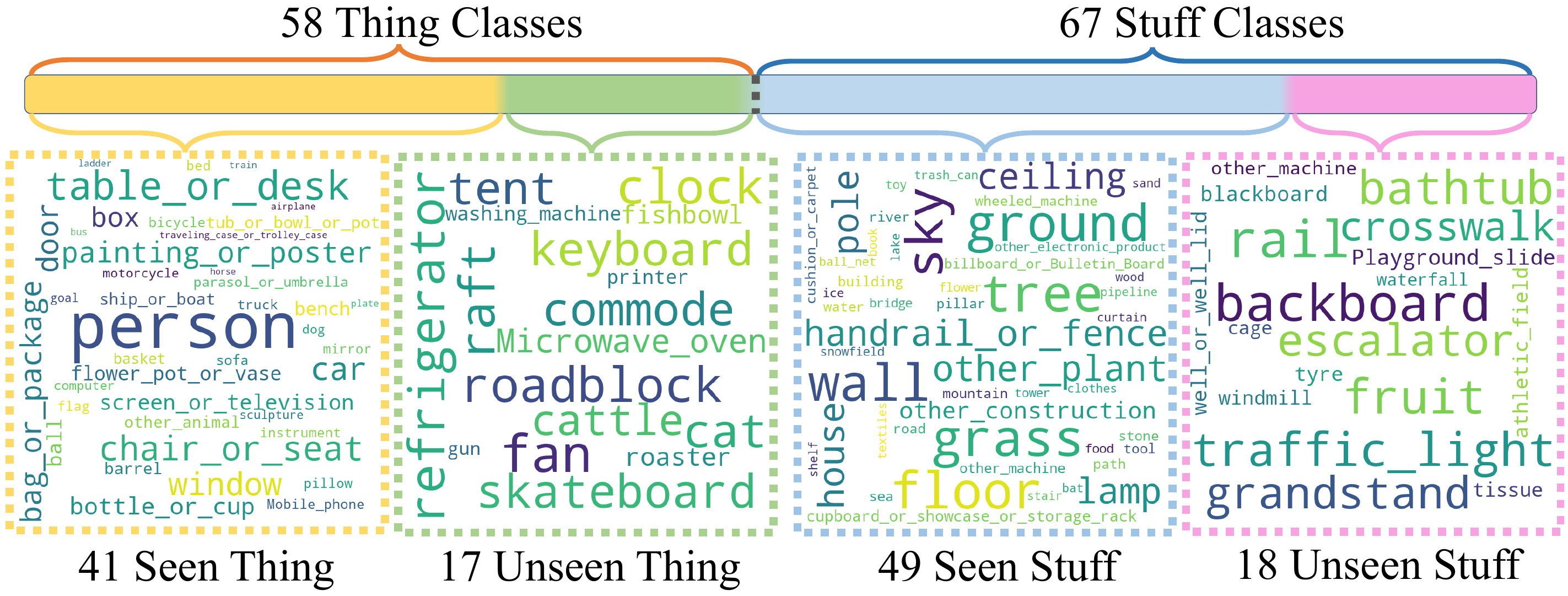}
\caption{Object classes and class subsets in VIPOSeg.}
\label{fig:class}
\end{figure}

\begin{figure}[t]
\subfigure[Object number distribution]{
\label{fig:obj}
\begin{minipage}[t]{0.64\columnwidth}
\includegraphics[width=\columnwidth]{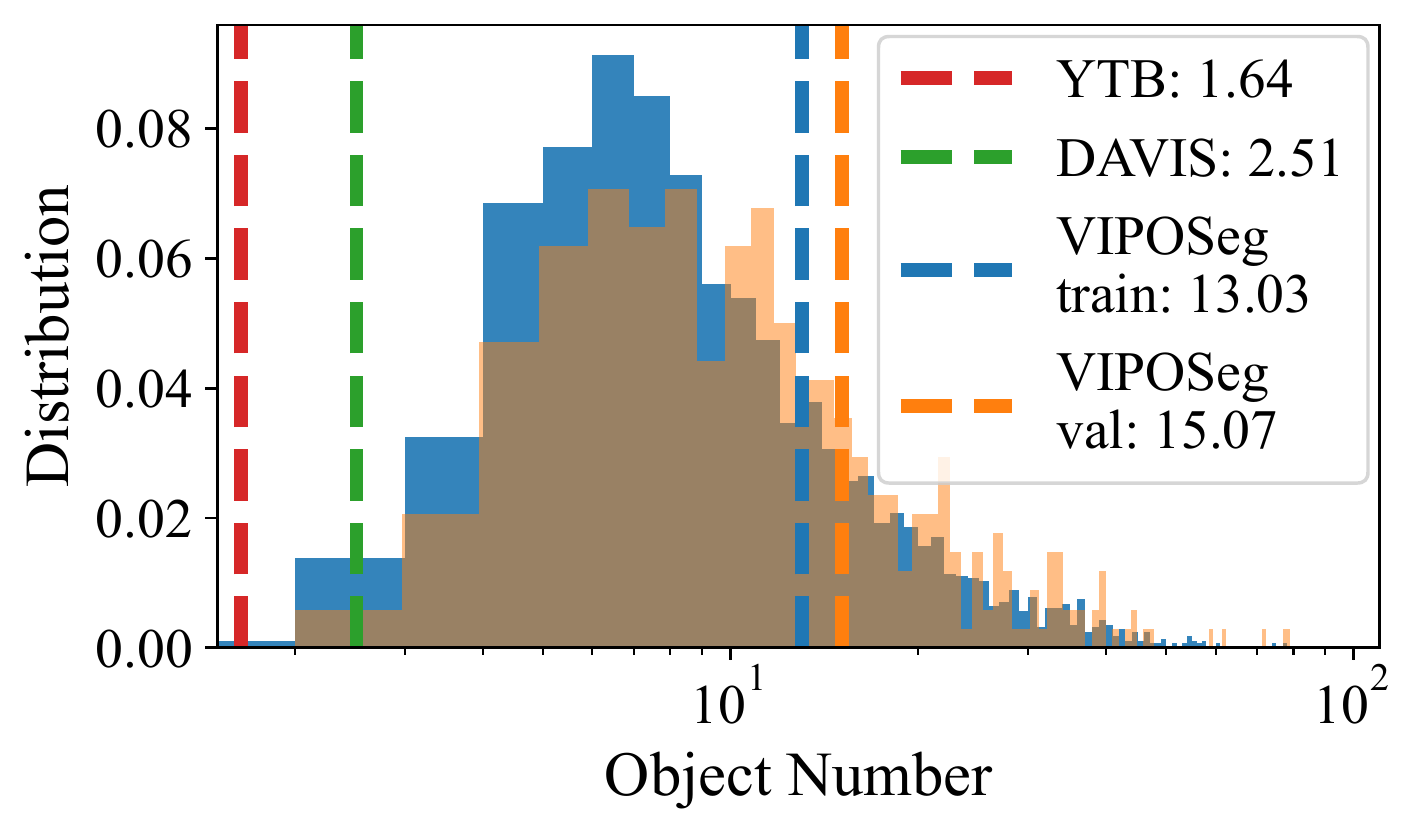}
\end{minipage}}
\subfigure[Scale ratios]{
\label{fig:scale}
\begin{minipage}[t]{0.33\columnwidth}
\includegraphics[width=\columnwidth]{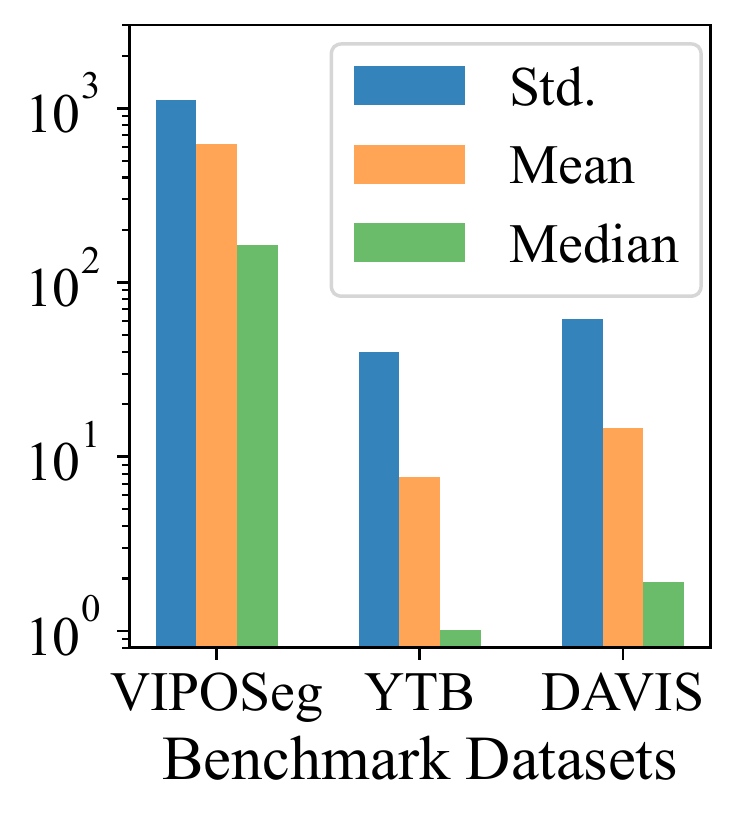}
\end{minipage}}
\caption{Comparison among VOS benchmark datasets including VIPOSeg, YouTube-VOS (YTB) and DAVIS. Figure (a) shows the object number distribution in VIPOSeg, as well as mean object numbers of other datasets. Figure (b) compares the distribution statistics of scale ratios in different datasets.}
\end{figure}

% \begin{figure}[t]
% \subfigure[Seen/unseen thing classes]{
% \label{fig:classa}
% \begin{minipage}[t]{0.48\columnwidth}
% \includegraphics[width=0.49\columnwidth]{pics/thing_seen.png}
% % \centering
% % \rule{3.5cm}{0.4pt}
% \includegraphics[width=0.49\columnwidth]{pics/thing_unseen.png}
% \end{minipage}}
% \subfigure[Seen/unseen stuff classes]{
% \label{fig:classb}
% \begin{minipage}[t]{0.48\columnwidth}
% \includegraphics[width=0.49\columnwidth]{pics/stuff_seen.png}
% % \centering
% % \rule{3.5cm}{0.4pt}
% \includegraphics[width=0.49\columnwidth]{pics/stuff_unseen.png}
% \end{minipage}}
% \caption{Four class subsets of VIPOSeg. Name sizes are drawn according to class frequencies.}
% \end{figure}

% \begin{figure}[t]
% \centering
% \includegraphics[width=1.0\columnwidth]{pics/noise-6.pdf}
% \caption{Annotation noise correction.}
% \label{fig:noise}
% \end{figure}

\subsection{Producing VIPOSeg}
Exhaustively annotating objects in images is extremely consuming, let alone in video frames. Fortunately, recently VIPSeg dataset \cite{miao2022large} provides 3536 videos annotated in a panoptic manner. It includes 124 classes consisting of 58 thing classes and 66 stuff classes. We adapt this dataset and build our VIPOSeg dataset based on it.

% Other datasets for VPS like Cityscape-VPS \cite{kim2020video}, KITTI-STEP and MOTChallenge-STEP \cite{weber2021step} are not considered because they focus on street-view scenes. Despite other datasets may also be adapted, VIPSeg is the best choice for panoptic VOS task. OVIS and UVO indeed contains denser object annotations than YouTube-VOS and DAVIS. However, OVIS only covers 25 classes and UVO is class-agnostic so the generalization ability of models on unseen classes is unknown. In addition, their annotations only contain the instances of thing classes but ignores stuff classes. Due to this reason, we still think the annotations are not exhaustive. 

\paragraph{Splitting dataset and classes.}
In terms of VIPSeg, the 3536 videos are split into 2,806/343/387 for training, validation and test. We only use training and validation sets in our VIPOSeg (3149 videos in total) because the annotations for test set are private. In order to add unseen classes to validation set, we re-split the videos into new training and validation set. We first sort 58 thing classes and 66 stuff classes respectively by frequency of occurrence. Next, we choose 17 thing classes and 17 stuff classes from the tail as unseen classes. We also split `other machine' into two classes, one for seen and another for unseen (detailed explanation is in supplementary material). Then, videos for validation set are selected by ensuring that enough objects in unseen classes should be included. Last but not least, we remove the annotations of unseen classes in training set. In summary, there are four subsets of 125 classes including 41/17 seen/unseen thing classes and 49/18 seen/unseen stuff classes (Figure \ref{fig:class}).

\paragraph{Creating and correcting annotations.}
In order to generate reference masks for VOS, we convert the panoptic annotations into object index format and then select the masks that appear the first time in each video as reference masks. To distinguish thing/stuff and seen/unseen classes, we also record the class mapping from object index to class index for each video. The class mapping enables us to calculate the evaluation metrics on seen/unseen and thing/stuff classes. Another problem is that the mask annotations in original VIPSeg are noisy, especially in the edges of objects. To ensure the correctness of evaluation, we manually recognize low-quality annotations and cleaned their noises in validation set.

% \paragraph{Cleaning annotations}
% The annotations in original VIPSeg \cite{miao2022large} of some video sequences are noisy, especially in the edges of objects. These noises may be taken as reference masks in VOS, which will lead to poor segmentation results and affect the reliability of evaluation. To ensure the correctness of reference masks, we recognize 24 videos whose reference masks are contaminated by noises in validation set. 741 frames in these 24 videos are manually checked, and 6561 noisy spots with 2549 objects involved are cleaned.
\paragraph{Settings of panoptic VOS.}
 The panoptic VOS task comes along with the rich and dense annotations. In panoptic VOS, models are trained with panoptic data. Besides, extra annotations indicating whether an object is thing or stuff are available in both training and test. Previous classic VOS, where only spatially sparse annotated data is used for training and test, can be regarded as a simplified version of panoptic VOS. Our method PAOT provides solutions for both panoptic/classic settings (Section \ref{sec:method}).

\subsection{Significance of VIPOSeg}
As a new benchmark dataset, VIPOSeg not only complements the deficiency of previous datasets but also surpasses them by a large margin in class diversity and object density. VIPOSeg has $4\times$ denser annotations, $20\times$ more videos than DAVIS, and $6\times$ denser annotations than YouTube-VOS (Table \ref{tab:datasets} and Figure \ref{fig:obj}). Denser annotations of panoptic scenes also includes objects on more diversified scales, with almost $30\times$ larger mean and $6400\times$ larger variance of scale ratios than YouTube-VOS (Figure \ref{fig:scale}). More importantly, VIPOSeg contains stuff classes which never appear in previous VOS datasets (Table \ref{tab:datasets}).

% More importantly, VIPOSeg will promote the development of panoptic VOS methods. VIPOSeg oriented panoptic VOS methods need to consider the generalization ability on unseen classes, the difference between thing and stuff classes and association among numerous objects. Therefore, panoptic VOS methods will show better robustness and efficiency in the real world applications than traditional VOS methods. Taking these into account, we believe panoptic VOS methods will play a important role in many applications, such as dense annotation propagation in video dataset labeling and crowd tracking in public places.

\begin{figure*}[t]
\centering
\includegraphics[width=0.98\textwidth]{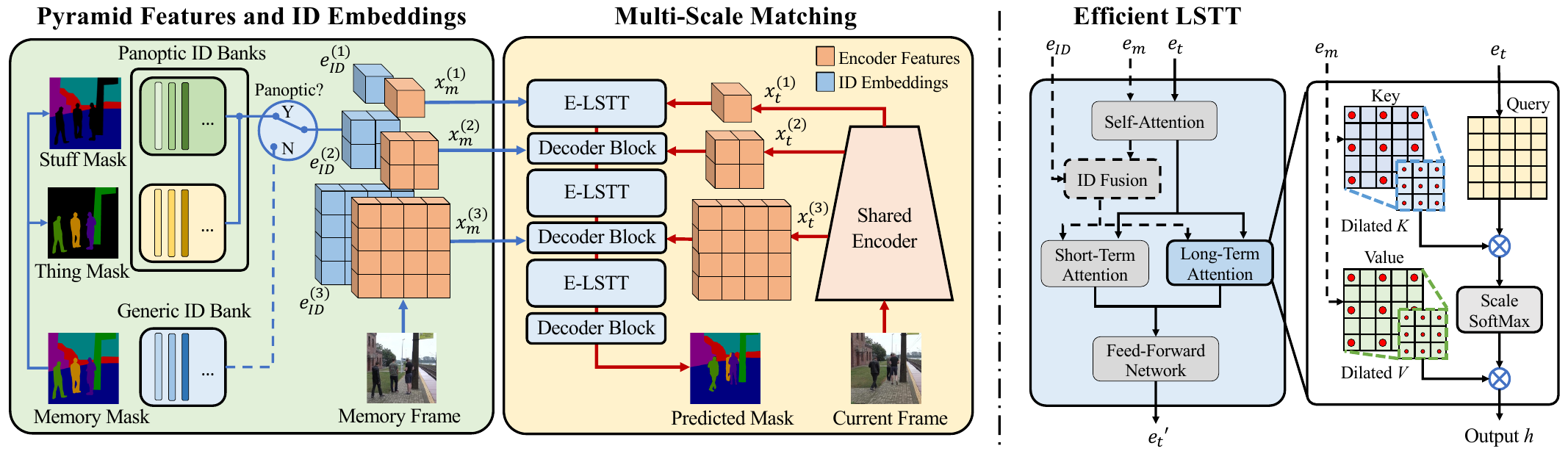}
\caption{Left part illustrates the generation of generic/panoptic ID embeddings and the pyramid architecture for multi-scale object matching. Right part shows the detailed structure of efficient long short-term transformer blocks and dilated long-short term attention in it.}
\label{fig:pyramid}
\end{figure*}

% \begin{figure}[t]
% \subfigure[Panoptic ID]{
% \label{fig:id1}
% \begin{minipage}[b]{0.6\columnwidth}
% \includegraphics[width=\columnwidth]{pics/pano_id-1.pdf}
% \end{minipage}}
% \subfigure[Generic ID]{
% \label{fig:id2}
% \begin{minipage}[t]{0.3\columnwidth}
% \includegraphics[width=\columnwidth]{pics/pano_id-2.pdf}
% \end{minipage}}
% \caption{Test.}
% \end{figure}

\begin{figure}[t]
\centering
\includegraphics[width=0.9\columnwidth]{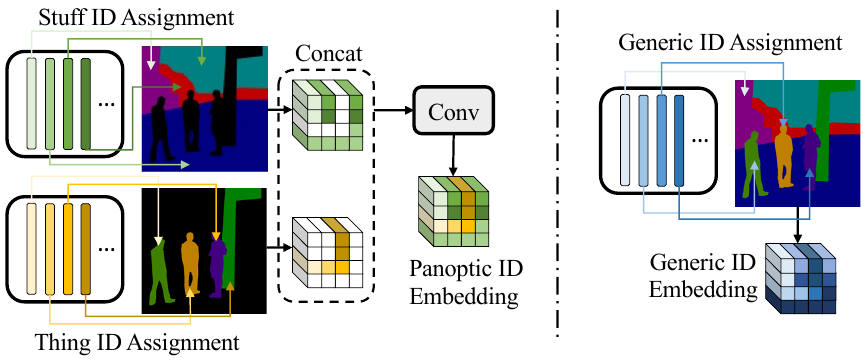}
\caption{Detailed illustration of panoptic ID embedding generation (left) and generic ID embedding generation (right).}
\label{fig:id}
\end{figure}

\subsection{Challenges in VIPOSeg}
With much denser object annotation and more diverse classes, challenges also emerge in the VIPOSeg dataset (Figure \ref{fig1}).
\paragraph{Motion and occlusion.}
Although previous datasets also include objects with motion and occlusion, they are not as challenging as VIPOSeg. The number of objects in VIPOSeg can be so large that the occlusion can be intractable.
\paragraph{Numerous objects.}
Another challenge that comes with the large number of objects is efficiency. Figure \ref{fig1} column two shows scenes with numerous objects and Figure \ref{fig:obj} shows the distribution of object number in VIPOSeg. VOS models will need more memory and be slower when the number of objects becomes larger. According to our experimental results in Table \ref{tab:fps}, CFBI+ \cite{yang2021collaborative} runs at 2 FPS and consumes over 30 GB memory when evaluated on VIPOSeg.
\paragraph{Various scales.}
% Since the scenes are exhaustively annotated, objects on all scales are included. First, scales can be hugely different between different classes. Stuff classes like `sky' and `ground' often takes a lot of space while thing classes like `traffic light' and `pole' are usually on a small scale. Second, the scales of objects  can have large difference because of the space perspective and zooming during photographing. Figure \ref{fig1} column three shows two examples and Figure \ref{fig:ms} gives more details.
Since the scenes are exhaustively annotated, objects on all scales are included. Figure \ref{fig:scale} shows the mean, median and standard deviation of the scale ratios in VOS benchmarks. The scale ratio is defined as the ratio of the pixel numbers of the largest and the smallest objects in a frame. The scale ratios of frames in VIPOSeg have much larger mean value and variance than previous benchmarks.
\paragraph{Unseen classes.} 
We deliberately wipe out the annotations of some classes in training set to make them unseen in validation set. Generalizing from seen to unseen is a common problem for most deep models. It is not easy to narrow the performance gap between seen and unseen.
\paragraph{Stuff classes.} 
Previous VOS datasets never contain stuff classes while VIPOSeg does. One may be curious about whether a VOS model can track the mask of flowing water like `sea wave' (Figure \ref{fig1} column three) and `waterfall' (Figure \ref{fig1} column four). The answer can be found in VIPOSeg.

\section{Method}
\label{sec:method}
In the face of above challenges, we develop a method Panoptic Object Association with Transformers (PAOT), which is not only designed for panoptic VOS but also compatible with classic VOS. PAOT consists of following designs,
1) For the motion and occlusion problem, we employ multi-object association transformers (AOT) \cite{yang2021associating} as the base framework.
2) For objects on various scales, a pyramid architecture is proposed to incorporate multi-scale features into the matching procedure.
3) For the thing/stuff objects in panoptic scenes, we decouple a single ID bank into two separate ID banks for thing and stuff to generate panoptic ID embeddings.
4) For the efficiency problem caused by numerous objects, an efficient version of long-short term transformers (E-LSTT) is proposed to balance performance and efficiency.

%-------------------------------------------------------------------------

% \begin{figure*}[t]

% \subfigure[AOT architecture]{
% \label{fig:aot}
% \begin{minipage}[t]{0.3\textwidth}
% \includegraphics[width=1\textwidth]{pics/AOT.pdf}
% \end{minipage}}
% \hfill
% \subfigure[Pyramid architecture with efficient LSTT]{
% \label{fig:PAOT}
% \begin{minipage}[t]{0.6\textwidth}
% \includegraphics[width=1\textwidth]{pics/PAOT.pdf}
% \end{minipage}}
% \caption{The figures above compares the architectures of AOT (a) and PAOT (b). The LSTT module is combined with the decoder to form a pyramid architecture in PAOT. The reduced attention in E-LSTT is also illustrated.}
% \end{figure*}

\subsection{Pyramid Architecture}
A pyramid architecture (Figure \ref{fig:pyramid}) is proposed in PAOT to perform matching on different scales. The scales are determined by the features $x^{(i)}$ from the encoder. For memory/reference frames who have masks, the mask information is encoded in ID embeddings $e_{ID}^{(i)}$ by assigning ID vectors in the ID bank. Each ID vector is corresponding to an object so the ID embedding contains information of all objects. The ID assignment can be regarded as a function which maps a one-hot label of multiple objects to a high dimensional embedding. Each scale $i$ has an individual ID bank to generate the ID embedding to maintain rich target information. The ID embedding is fused with the memory frame embedding $e_m^{(i)}$ as key and value, waiting for the query of later frames.

For a current frame without a mask, the E-LSTT module is responsible for performing matching between the embeddings of the current frame $e_t^{(i)}$ and memory/reference frames $e_m^{(i)}$.  Next, the decoder block is able to decode the matching information and incorporate the features on the larger scale $x_t^{(i+1)}$. The matching and decoding process is in a recursive manner from the current scale to the next scale,

\begin{align*}\label{eq:pareto mle2}
e_t^{(i)'} &= T^{(i)}_E(e_t^{(i)}, e_m^{(i)}, e_{ID}^{(i)}),\\
e_t^{(i+1)} &= R^{(i)}(s^+(e_t^{(i)'})+ x_t^{(i+1)}),
\end{align*}
where $T^{(i)}_E(\cdot)$ is the E-LSTT module, $s^+(\cdot)$ is the up-sampling function and $R^{(i)}(\cdot)$ is the decoder block (implemented as residual convolutional blocks \cite{he2016deep}). 
%Note there is no $x^{(i+1)}$ for the last scale. Residual blocks are omitted in Figure \ref{fig:pyramid} for simplicity.

% The pyramid architecture has two advantages. First, it fully utilizes the feature maps on different scales. AOT only uses the feature maps on the smallest scale to perform matching, while the pyramid architecture enables feature maps on multiple scales to be involved in matching. Second, it deepens the whole model and the model capacity also extends along with the depth. In PAOT, each stage includes several LSTT blocks and the information aggregated by the previous stage can be accumulated and reused in the current stage. As a result, the number of LSTT layers increases and the model continues to gain performance improvement.

\subsection{Generation of Panoptic ID Embeddings}
For panoptic VOS, we generate panoptic ID embedding from thing and stuff mask on each scale (Figure \ref{fig:pyramid}, \ref{fig:id}). Previous VOS datasets and methods only consider the countable thing objects but omit stuff objects. Although thing objects and stuff objects can be treated equally in a unified manner in classic VOS methods, the difference between stuff and thing should not be ignored. Considering this, we decouple the ID bank into two separate ID banks for thing and stuff objects respectively. We aim to obtain more discriminative ID embeddings for thing objects while more generic ID embeddings for stuff objects, especially unseen stuff objects. 

The label of a frame $y$ is first decomposed into thing label $y_{th}$ and stuff label $y_{st}$. The thing objects are assigned with ID vectors from thing ID bank and stuff objects are assigned wtih ID vectors from stuff ID bank. Last, the thing and stuff ID embeddings are concatenated and fed into the aggregation module (implemented as convolutional layers) to obtain the panoptic ID embedding on scale $i$,

\begin{equation*}
e_{ID}^{(i)} = Conv(Cat(ID^{(i)}_{th}(y_{th}),ID^{(i)}_{st}(y_{st}))).
\end{equation*}

\subsection{Efficient Long Short-Term Transformers}
% Directly using LSTT in the pyramid structure causes two problems. First, the increase in the number of transformer layers leads to the increase in the amount of calculation. Second, the use of large scale feature maps in attention causes great demand for the memory space. Taking these problems into account, we design a more efficient structure for the long short-term transformer (E-LSTT), illustrated in Figure \ref{fig:elstt}.

% Third, the introduction of large scale feature maps undermines the learning of features on the small scale. 

Long-short term transformers (LSTT) are proposed in AOT \cite{yang2021associating} for object matching. Directly using LSTT in the pyramid structure causes efficiency problem due to the larger scales, and the problem will become more serious in panoptic scenes due to numerous objects.

The long-term attention dominates the computational cost of LSTT because the attention may involve multiple memory frames. In order to cut down the computational cost, we use single-head rather than multi-head attention for the long-term memory. Inspired by \cite{wang2021pyramid}, we further apply the dilated attention where the key and value are down-sampled in the long-term attention on large scales (Figure \ref{fig:pyramid}). More details can be found in supplementary material. 

\begin{figure}[t]
\centering
\includegraphics[width=0.98\columnwidth]{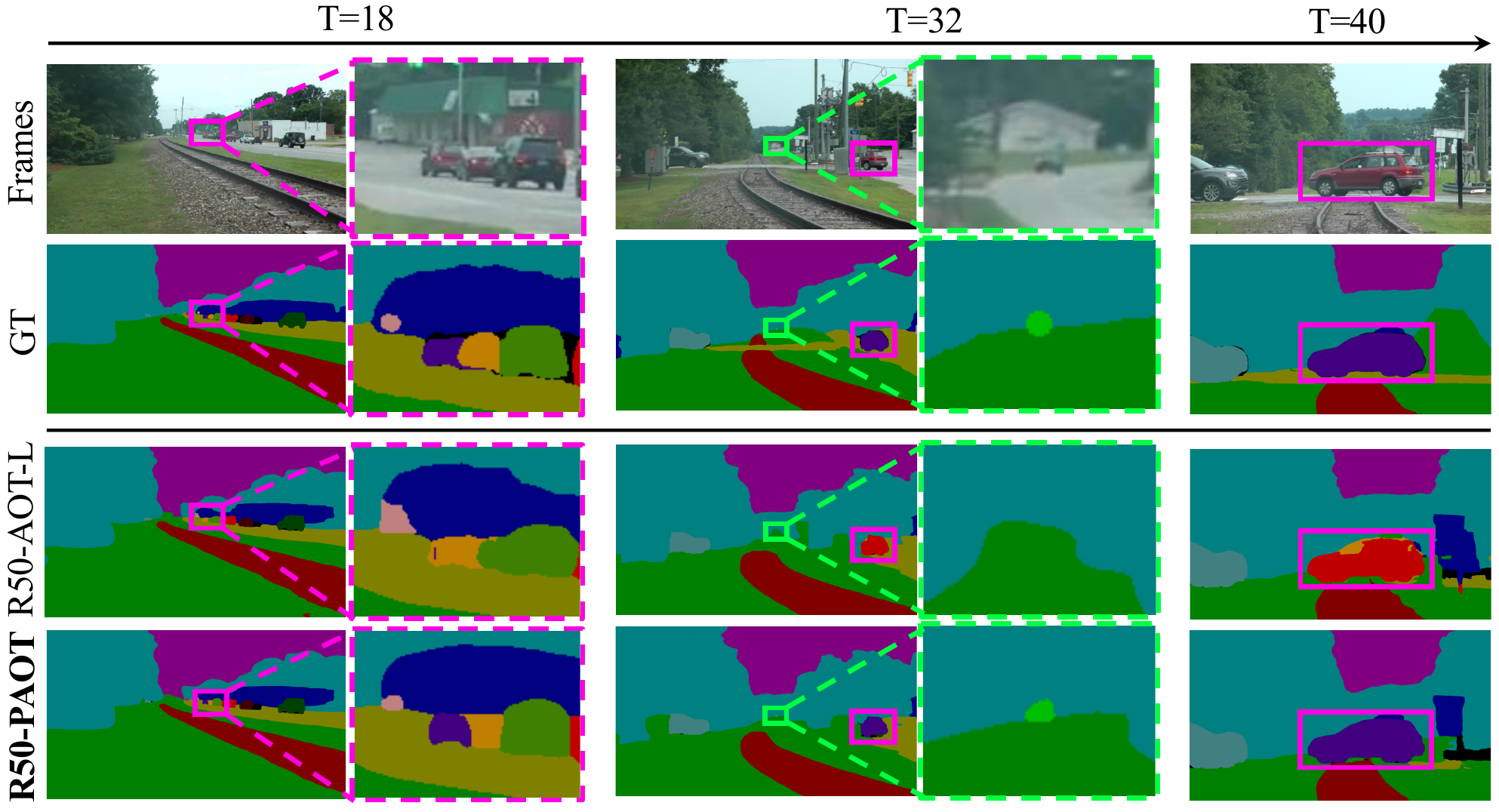}
\caption{Comparison between AOT and PAOT models on a video sequence with objects on various scales. Small objects in boxes are enlarged for better viewing.}
\label{fig:ms}
\end{figure}

\section{Experiment}
\subsection{Implementation Details}
\paragraph{Model settings.}
The encoder backbones of PAOT models are chosen in ResNet-50 \cite{he2016deep} and Swin Transformer-Base \cite{liu2021swin}. As for multi-scale object matching, we set E-LSTT in four scales $16\times,16\times,8\times,4\times$ to be 2,1,1,0 layers respectively (4 layers in total). It should be noted that we do not use the $4\times$ scale feature maps for object matching but only for decoding considering the computational burden, and instead duplicate the $16\times$ features twice.

\paragraph{Training procedure.}
The training procedure consists of two steps: (1) pre-training on the synthetic video sequences generated by static image datasets \cite{everingham2010pascal,lin2014microsoft,cheng2014global,shi2015hierarchical,hariharan2011semantic} by randomly applying multiple image augmentations \cite{oh2018fast}. (2) main training on the real video sequences by randomly applying video augmentations \cite{yang2020collaborative}. The datasets for training include DAVIS 2017 (D) \cite{pont20172017}, YouTube-VOS 2019 (Y) \cite{xu2018youtube} and our VIPOSeg (V). Models pre-trained with BL-30K \cite{cheng2021modular} are marked with $\ast$ (for STCN \cite{cheng2021rethinking}).

During training, we use 4 Nvidia Tesla A100 GPUs, and the batch size is 16. For pre-training, we use an initial learning rate of $4 \times 10^{-4}$ for 100,000 steps. For main training, the initial learning rate is $2 \times 10^{-4}$, and the training steps are 100,000. The learning rate gradually decays to $1 \times 10^{-5}$ in a polynomial manner \cite{yang2020collaborative}.

\paragraph{Task settings.}
 For panoptic setting, V is used for training and evaluation. PAOT models with panoptic ID are marked with Pano-ID, otherwise generic ID is used. Note that PAOT with generic ID is compatible with classic VOS. For classic setting, Y+D are used for training and evaluation. Training with Y+D+V is mainly for classic setting and V is regarded as auxiliary data.

\subsection{Evaluation Results on VIPOSeg}

% \usepackage{booktabs}
% \usepackage{graphicx}
\begin{table*}[]
\centering
\scriptsize
\setlength{\tabcolsep}{2pt}
\renewcommand{\arraystretch}{0.8}
\resizebox{\textwidth}{!}{%
\begin{tabular}{@{}lcc|cccc|cccc|cccc|c@{}}
\toprule
                                         & \multicolumn{1}{l}{} & \multicolumn{1}{l|}{} & \multicolumn{4}{c|}{Average IoU}                                                                & \multicolumn{4}{c|}{Mask IoU}                                 & \multicolumn{4}{c|}{Boundary IoU}                             & \multicolumn{1}{l}{} \\ \midrule
\multicolumn{3}{c|}{VIPOSeg Validation}                                                 & \multicolumn{2}{c}{Seen/Unseen}   & \multicolumn{2}{c|}{Thing/Stuff}                            & \multicolumn{2}{c}{Thing}     & \multicolumn{2}{c|}{Stuff}    & \multicolumn{2}{c}{Thing}     & \multicolumn{2}{c|}{Stuff}    & \multicolumn{1}{l}{} \\ \midrule
Methods                                  & Training             & $\mathcal{G}$         & $\mathcal{G}_s$ & $\mathcal{G}_u$ & \multicolumn{1}{l}{$\mathcal{G}^{th}$} & $\mathcal{G}^{sf}$ & $M^{th}_{s}$  & $M^{th}_{u}$  & $M^{sf}_{s}$  & $M^{sf}_{u}$  & $B^{th}_{s}$  & $B^{th}_{u}$  & $B^{sf}_{s}$  & $B^{sf}_{u}$  & $\lambda$            \\ \midrule
CFBI+ \cite{yang2021collaborative}       & Y+D                  & 72.1                  & 73.0            & 71.3            & 68.3                                   & 76.0               & 69.6          & 69.4          & 80.4          & 77.3          & 67.7          & 66.7          & 74.4          & 71.7          & 1.42                 \\
STCN \cite{cheng2021rethinking}          & Y+D                  & 72.4                  & 73.8            & 71.1            & 68.4                                   & 76.5               & 70.9          & 68.3          & 80.8          & 78.3          & 69.0          & 65.5          & 74.5          & 72.3          & 1.03                 \\
STCN$^{\ast}$ \cite{cheng2021rethinking} & Y+D                  & 72.5                  & 73.6            & 71.4            & 69.0                                   & 76.0               & 71.2          & 69.1          & 80.1          & 78.2          & 69.4          & 66.4          & 73.9          & 72.0          & 1.08                 \\
R50-AOT-L  \cite{yang2021associating}    & Y+D                  & 73.7                  & 74.8            & 72.6            & 72.1                                   & 75.4               & 73.9          & 72.6          & 79.7          & 77.2          & 72.1          & 69.8          & 73.6          & 70.9          & 0.88                 \\
SwinB-AOT-L  \cite{yang2021associating}  & Y+D                  & 73.3                  & 74.5            & 72.2            & 72.0                                   & 74.6               & 74.4          & 71.9          & 78.8          & 76.9          & 72.5          & 69.3          & 72.2          & 70.6          & 0.92                 \\ \midrule
\textbf{R50-PAOT}                        & Y+D                  & \textbf{75.4}         & \textbf{76.5}   & 74.3            & 74.2                                   & \textbf{76.5}      & 76.0          & 74.7          & \textbf{80.7} & \textbf{78.3} & 74.2          & 72.0          & \textbf{74.9} & \textbf{72.2} & \textbf{0.84}        \\
\textbf{SwinB-PAOT}                      & Y+D                  & 75.3                  & 76.3            & \textbf{74.4}   & \textbf{74.7}                          & 76.0               & \textbf{76.4} & \textbf{75.0} & 80.0          & 77.9          & \textbf{74.7} & \textbf{72.5} & 74.1          & 72.1          & 0.87                 \\ \midrule\midrule
R50-AOT-L \cite{yang2021associating}     & V                    & 76.4                  & 78.0            & 74.8            & 74.2                                   & 78.6               & 76.8          & 73.8          & 82.9          & 80.0          & 75.0          & 71.2          & 77.3          & 74.1          & 0.78                 \\
\textbf{R50-PAOT}                        & V                    & 77.5                  & \textbf{79.1}            & 75.8            & 75.9                                   & 79.0               & \textbf{78.2}          & 75.7          & \textbf{83.6}          & 79.9          & \textbf{76.5}          & 73.2          & \textbf{78.2}          & 74.4          & 0.77                 \\
\textbf{R50-PAOT (Pano-ID)}                        & V                    & \textbf{77.9}         & 79.0            & \textbf{76.8}   & \textbf{76.0}                                   & \textbf{79.8}      & 78.1          & \textbf{76.0}          & 83.3          & \textbf{81.8} & 76.4          & \textbf{73.3}          & 77.9          & \textbf{75.9} & \textbf{0.76} \\\midrule
\textbf{SwinB-PAOT}                      & V                    & 78.0         & 79.5   & 76.5   & \textbf{76.4}                          & 79.6      & \textbf{78.8} & \textbf{76.0} & 83.7 & 80.8 & \textbf{77.2} & 73.7 & 78.3 & 75.5 & 0.70        \\ 
\textbf{SwinB-PAOT (Pano-ID)}                      & V                    & \textbf{78.2}         & \textbf{79.5}            & \textbf{76.9}   & 76.3                                   & \textbf{80.1}      & 78.6          & 75.9          & \textbf{83.9}          & 81.7 & 76.9          & \textbf{73.7}          & \textbf{78.5}          & \textbf{76.2} & \textbf{0.70}  \\ \midrule\midrule
R50-AOT-L \cite{yang2021associating}     & Y+D+V                & 76.5                  & 77.9            & 75.0            & 74.3                                   & 78.6               & 76.7          & 74.1          & 82.8          & 80.2          & 74.9          & 71.6          & 77.2          & 74.2          & 0.80                 \\
\textbf{R50-PAOT}                        & Y+D+V                & 77.4                  & 78.4            & 76.4            & 75.9                                   & 78.8               & 77.5          & \textbf{76.6} & 82.9          & 80.3          & 75.8          & \textbf{73.9} & 77.5          & 74.7          & 0.79                 \\
\textbf{SwinB-PAOT}                      & Y+D+V                & \textbf{77.9}         & \textbf{79.3}   & \textbf{76.5}   & \textbf{76.3}                          & \textbf{79.5}      & \textbf{78.8} & 75.8          & \textbf{83.3} & \textbf{81.2} & \textbf{77.2} & 73.5          & \textbf{77.8} & \textbf{75.7} & \textbf{0.73}        \\ \bottomrule
\end{tabular}%
}
\caption{Evaluation results on VIPOSeg validation set. Training datasets include YouTube-VOS (Y), DAVIS (D) and VIPOSeg (V). $\ast$ denotes that models are pre-trained with BL-30K. $\lambda$ is the decay constant.}
\label{tab:viposeg}
\end{table*}

% VIPOSeg is our new benchmark that has comparable scale with YouTube-VOS but contains panoptic annotations with $6\times$ denser objects than YouTube-VOS. The validation set has 384 video sequences with 58 thing classes and 67 stuff classes. The videos in validation set with 124 classes cover a variety of scenes in real world. 17 of 58 thing classes and 18 of 67 stuff classes are unseen in the training set.

\paragraph{Evaluation metrics.} For a new benchmark, it is crucial to choose proper metrics to evaluate the performance. 
%We follow previous benchmarks and further make some improvement. For mask evaluation, we keep Jaccord score (i.e. the mask IoU) and substitute boundary IoU \cite{cheng2021boundary} for boundary F-measure. According to the paper \cite{cheng2021boundary}, F-measure ignores small contour misalignment that can be attributed to ambiguity. In panoptic scenes, various sizes of objects are included. Boundary IoU is more likely to take more objects into account while F-measure may ignore small objects. In summary, 
We set eight separate metrics including \textbf{four mask IoUs} for seen/unseen thing/stuff ($\mathcal{M}^{th}_s$/$\mathcal{M}^{th}_u$/$\mathcal{M}^{sf}_s$/$\mathcal{M}^{sf}_u$), and \textbf{four boundary IoUs} \cite{cheng2021boundary} for seen/unseen thing/stuff  ($\mathcal{B}^{th}_s$/$\mathcal{B}^{th}_u$/$\mathcal{B}^{sf}_s$/$\mathcal{B}^{sf}_u$) respectively. The overall performance $\mathcal{G}$ is the average of these eight metrics. Moreover, four average metrics are calculated to indicate the average performance on thing/stuff ($\mathcal{G}^{th}$/$\mathcal{G}^{sf}$) and seen/unseen ($\mathcal{G}_s$/$\mathcal{G}_u$). The results with these metrics can be found in Table \ref{tab:viposeg}. Except for these standard metrics, there is also a special metric on VIPOSeg, the decay constant $\lambda$. It is in charge of evaluating the robustness of models in crowded scenes. More details can be found in later \textbf{Crowd decay} section.

\paragraph{Panoptic setting.} We train AOT \cite{yang2021associating} and PAOT models with VIPOSeg (V) as in panoptic VOS. The evaluation results are in middle of Table \ref{tab:viposeg}. Both the pyramid architecture and panoptic IDs in PAOT are beneficial to panoptic scenes. First, our PAOT model with generic IDs surpasses AOT by 1.1\% with the same R-50 backbone, which shows the improvement of the pyramid architecture. Second, the PAOT models with panoptic IDs have higher overall performance than PAOT models with generic IDs. Their difference is mainly on the metrics of unseen and stuff. R50-PAOT (Pano-ID) have around  \textbf{2\% higher mask IoU $M_{u}^{sf}$ and  1.5\% higher boundary IoU $B_{u}^{sf}$ on unseen stuff} objects than generic R50-PAOT. Therefore, decoupling the ID bank into thing and stuff is beneficial to learn more robust stuff ID vectors which generalize better on unseen objects.

\begin{figure*}[t]
\centering
\includegraphics[width=0.49\textwidth]{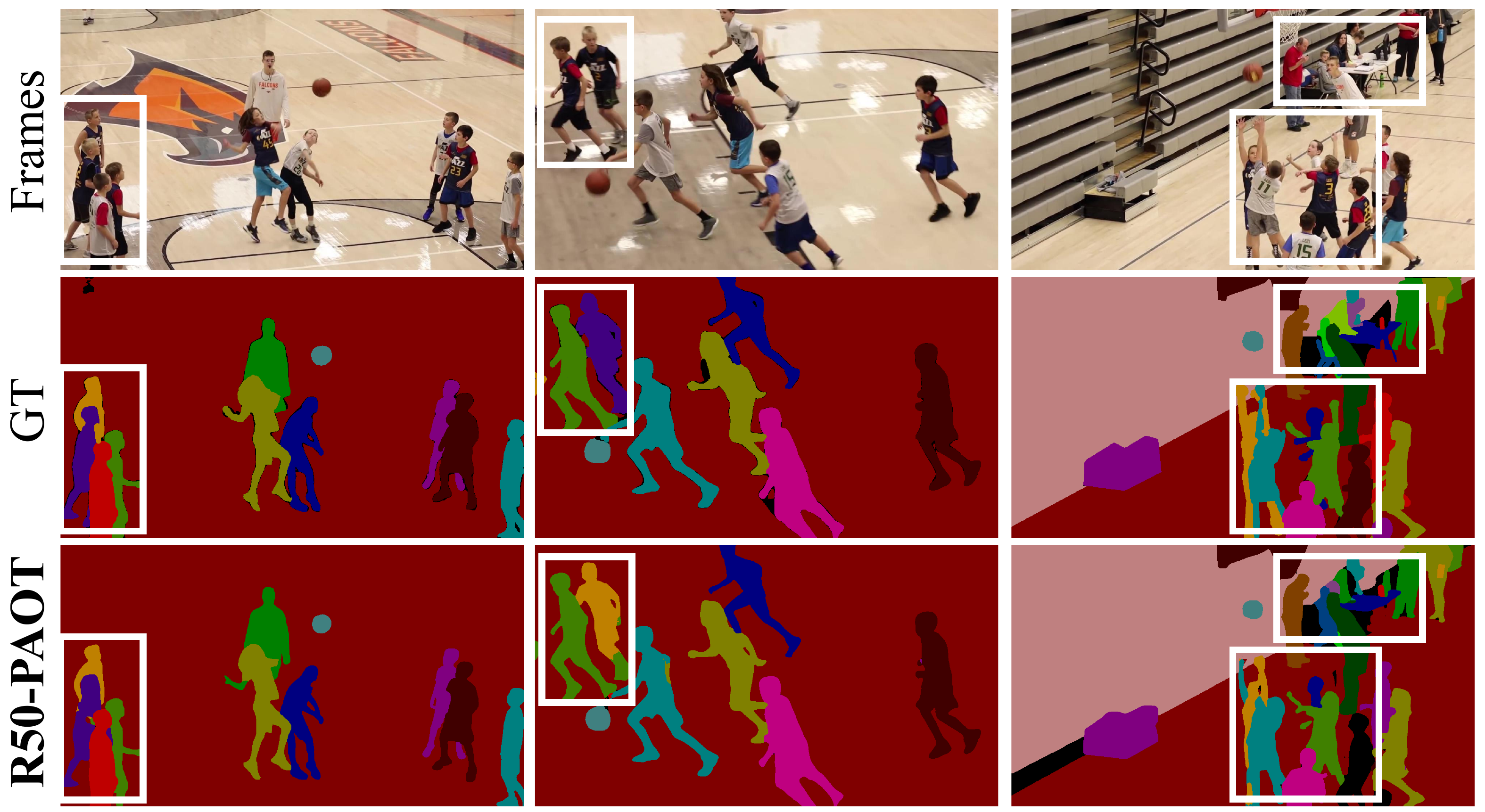}
\includegraphics[width=0.49\textwidth]{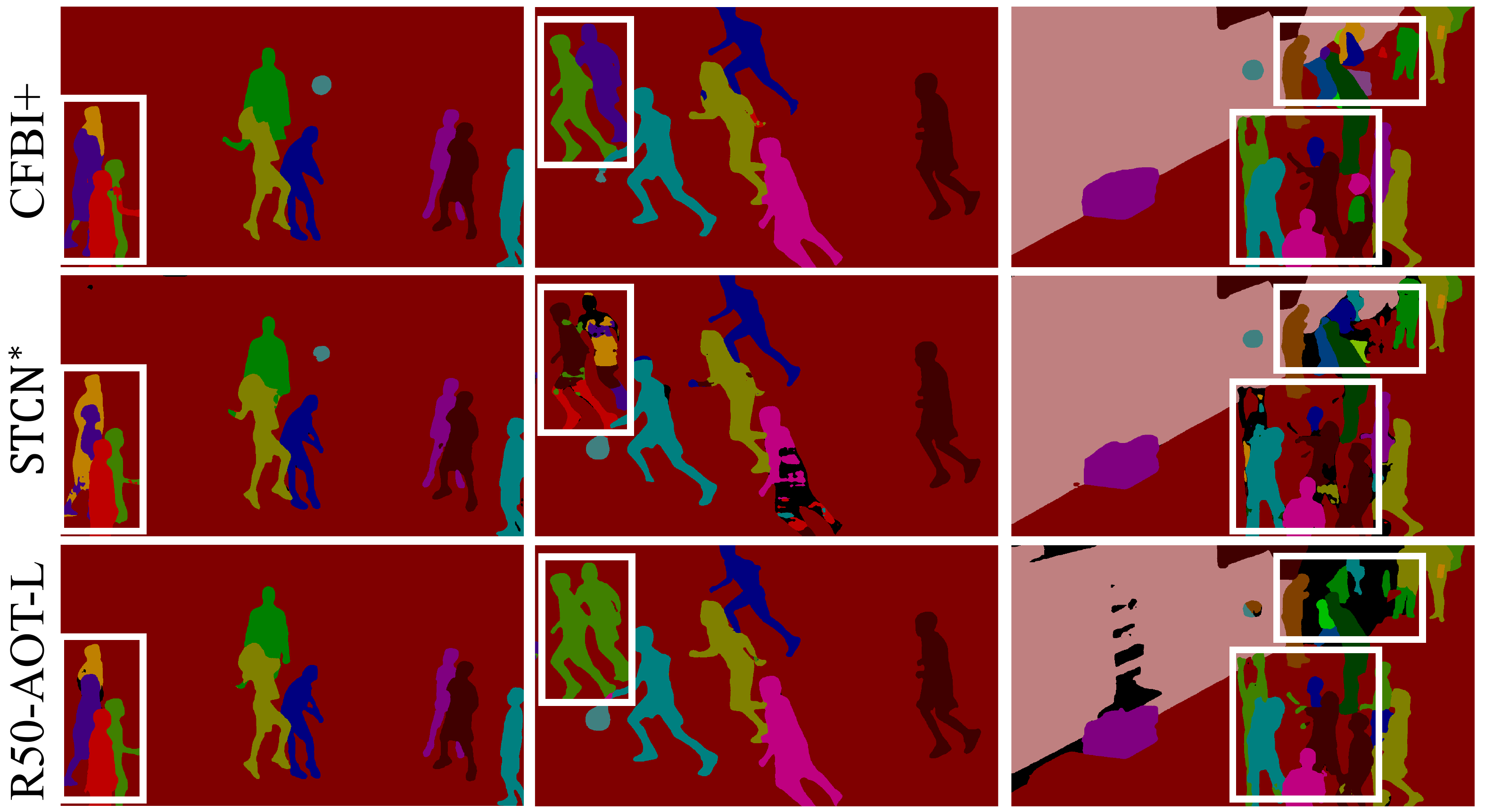}
\caption{Qualitative results of different methods evaluated on VIPOSeg validation set. The scene is a basketball contest and includes multiple players moving fast and occluding each other. Difficult areas are marked with boxes. }
\label{fig:qua}
\end{figure*}

\paragraph{Classic setting.} We test several representative methods including CFBI+ \cite{yang2021collaborative}, STCN \cite{cheng2021rethinking}, AOT \cite{yang2021associating} and our PAOT on VIPOSeg validation set. The evaluation results are in top of Table \ref{tab:viposeg}. These models are trained with Y+D. First, the overall IoU scores of previous methods are around 73.0. Compared with them, our PAOT models are above 75.0, which \textbf{surpass previous methods by over 2\%}. Qualitative results of these methods are in Figure \ref{fig:qua}. Second, previous methods like CFBI+ and STCN perform poorly on thing IoU $\mathcal{G}^{th}$. By contrast, multi-object association based methods like AOT and PAOT improve thing IoU a lot because the simultaneous multi-object propagation with ID mechanism is capable of modeling multi-object relationship such as occlusion.

\paragraph{Boosting performance by panoptic training.}
The overall IoU of AOT or PAOT \textbf{rises around 3\%} after replacing the training data from Y+D to V. There is a huge gap between models trained with and without VIPOSeg. The VIPOSeg training set enables the models to learn panoptic object association and to generalize in more complex scenes and classes. Besides, panoptic training data also benefits VOS models on previous classic VOS benchmarks, as shown in Table \ref{tab:ytb}.

\begin{figure}[t]
% \centering
\includegraphics[width=0.98\columnwidth]{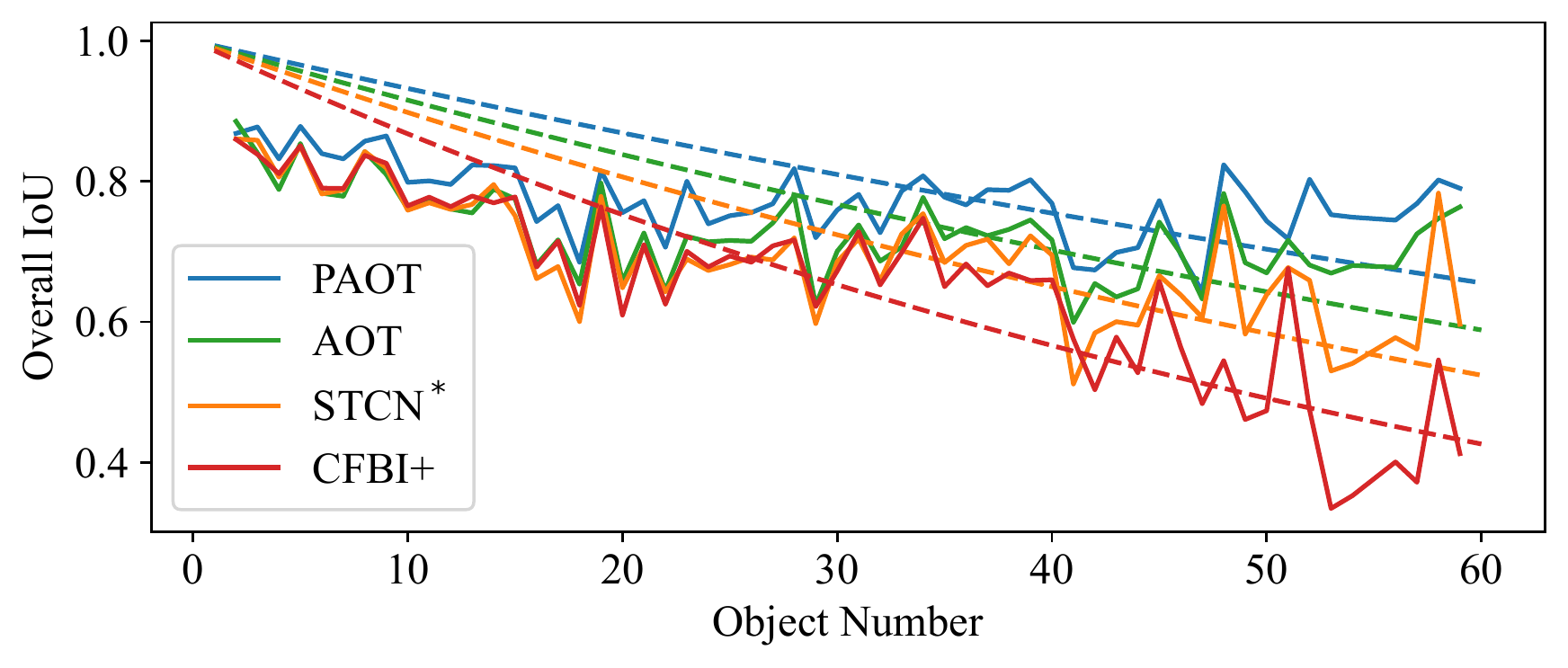}
\caption{Crowd decay of different methods on VIPOSeg.}
\label{fig:decay}
\end{figure}

\paragraph{Crowd decay.}
Dense annotations in VIPOSeg enable us to evaluate the performance of models under scenes with different amounts of objects. Here we present the crowd decay evaluation. 
% First, given an object number $n$, we collect all frames which contain $n$ objects and calculate average IoU $\mathcal{G}$. Then, 
We model the problem as exponential decay, $\mathcal{G}(n)=e^{-\lambda n/s}$ where $s=100$ is a scaling factor and $\lambda$ is the decay constant that reflects how fast the performance $\mathcal{G}$ drops when object number $n$ increases. The IoU for each object number $n$ is collected to estimate $\lambda$ by least square. We show the decay constants for different methods and models in Table \ref{tab:viposeg} and plot the decay curves in Figure \ref{fig:decay}. The results show that multi-object association methods (AOT, PAOT are around 0.8) have lower decay constants than other methods (CFBI+, STCN are above 1.0). SwinB-PAOT trained with VIPOSeg achieves the lowest decay constant 0.70, which means it deals with crowded scenes better than other models.

\paragraph{Speed and memory.} For all methods evaluated on VIPOSeg, we record the FPS and maximal memory space they consume during evaluation, which can be found in Table \ref{tab:fps}. The measure of FPS and memory is on Nvidia Tesla A100 GPU. CFBI+ runs at 2 FPS while STCN and AOT are at around 10 FPS. This fact shows VIPOSeg benchmark is very challenging in model efficiency. STCN runs faster with more memory while AOT and PAOT run slightly slower with less memory. However, all of these models demond over 11 GB memory, which leaves a large space for further improvement. A larger ID capacity and better memory strategy may help with the efficiency problem.

% \usepackage{booktabs}
% \usepackage{graphicx}
\begin{table}[]
\centering
\small
\scriptsize
\setlength{\tabcolsep}{2pt}
\renewcommand{\arraystretch}{0.8}
\resizebox{\columnwidth}{!}{%
\begin{tabular}{@{}lc|c|cc@{}}
\toprule
Methods                                      & IDs  & $\mathcal{G}$ & Total FPS      & Memory/GB      \\ \midrule
CFBI+ \cite{yang2021collaborative}         & -    & 72.1          & 2.01           & 33.13          \\
STCN  \cite{cheng2021rethinking}           & -    & 72.5          & \textbf{11.60} & 14.17          \\
R50-AOT-L \cite{yang2021associating}       & 10   & 73.7          & 11.30          & 12.35          \\
SwinB-AOT-L \cite{yang2021associating}     & 10   & 73.3          & 9.13           & 12.21          \\ \midrule\midrule
\textbf{R50-PAOT$^{\ddagger}$}             & 10   & 77.5          & 10.45          & 11.04 \\
\textbf{SwinB-PAOT$^{\ddagger}$}           & 10   & 78.0          & 8.35           & 11.18          \\ \midrule
\textbf{R50-PAOT$^{\ddagger}$ (Pano-ID)}   & 10+5 & \textbf{77.9} & 11.23          & 10.58          \\
\textbf{SwinB-PAOT$^{\ddagger}$ (Pano-ID)} & 10+5 & \textbf{78.2} & 8.48           & 10.72          \\ \midrule
R50-PAOT$^{\ddagger}$                      & 15   & 77.4          & 12.60          & 9.67           \\
R50-PAOT$^{\ddagger}$                      & 20   & 77.4          & 13.32          & 8.34           \\
R50-PAOT$^{\ddagger}$                      & 30   & 76.5          & \textbf{14.16} & \textbf{7.96}  \\ \bottomrule
\end{tabular}%
}
\caption{Speed and memory consumption of different methods on VIPOSeg validation set. $\ddagger$ denotes models trained with V rather than Y+D. Memory is the maximal GPU memory used by the method.}
\label{tab:fps}
\end{table}

\begin{table}[]
\centering
\scriptsize
\setlength{\tabcolsep}{2pt}
\renewcommand{\arraystretch}{0.8}
\resizebox{\columnwidth}{!}{%
\begin{tabular}{@{}l|c|c|c|c|c@{}}
\toprule
Methods                                  & Training & Y19                   & D17           & D17-T         & D16           \\ \midrule
CFBI+ \cite{yang2021collaborative}       & Y+D      & 82.6                  & 82.9          & 74.8          & 89.9          \\
HMMN \cite{seong2021hierarchical}        & Y+D      & 82.5                  & 84.7          & 78.6          & 90.8          \\
STCN \cite{cheng2021rethinking}          & Y+D      & 82.7$^\star$          & 85.4          & 76.1          & 91.6          \\
STCN$^{\ast}$ \cite{cheng2021rethinking} & Y+D      & 84.2$^\star$          & 85.3          & 79.9          & 91.7          \\
RPCM \cite{xu2022reliable}               & Y+D      & 83.9                  & 83.7          & 79.2          & 91.5          \\
R50-AOT-L \cite{yang2021associating}     & Y+D      & 85.3$^\star$          & 84.9          & 79.6          & 91.1          \\
SwinB-AOT-L \cite{yang2021associating}   & Y+D      & 85.3$^\star$          & 85.4          & 81.2          & 92.0          \\ \midrule
\textbf{R50-PAOT}                        & Y+D      & 85.9$^\star$          & 85.3          & 81.0          & 92.2          \\
\textbf{R50-PAOT}                        & Y+D+V    & \textbf{86.1$^\star$} & \textbf{86.0} & \textbf{82.1} & \textbf{92.5} \\ \midrule
\textbf{SwinB-PAOT}                      & Y+D      & 86.4$^\star$          & 86.2          & \textbf{84.0} & \textbf{93.8} \\
\textbf{SwinB-PAOT}                      & Y+D+V    & \textbf{86.9$^\star$} & \textbf{87.0} & 83.6          & 93.3          \\ \bottomrule
\end{tabular}%
}
\caption{Evaluation results on YouTube-VOS 2019 validation (Y19), DAVIS 2016/2017 validation (D16/D17) and 2017 test (D17-T). $\star$ for Y19 denotes testing using all frames. $^\ast$denotes models pre-trained with BL-30K.}
\label{tab:ytb}
\end{table}

% \subsubsection{Qualitative results}
% We present the predicted masks of different methods tested on a video sequence `880\_z1H\_RAzEnYs' in VIPOSeg validation set in Figure \ref{fig:qua}. The video sequence is recorded in a basketball contest and there are over 10 people moving fast in the scene. Boxes suggest difficult parts in the scene. Despite a few failure cases, our PAOT model outperforms other methods.

\subsection{Results on YouTube-VOS and DAVIS}

The evaluation results on YouTube-VOS \cite{xu2018youtube} and DAVIS \cite{perazzi2016benchmark,pont20172017} are listed in Table \ref{tab:ytb}. More detailed tables can be found in supplementary material. For models trained with Y+D, our PAOT model with Swin Transformer-Base backbone achieves SOTA performance on all benchmarks. Adding VIPOSeg to training can further boost performance.

\section{Ablation Study and Discussion}
% In this section, we provide ablation experimental results on PAOT design.
\paragraph{Capacity of ID banks.}
% According to AOT \cite{yang2021associating}, the capacity of the ID bank can influence the performance. Here 
The capacity of ID banks is a trade-off between efficiency and performance. The results are in Table \ref{tab:fps}. When the ID number increases, the performance drops while the speed rises and memory consumption decreases. Training more IDs results in less training data for each ID on average, and IDs with poorer generalization ability may affect the performance. For the classic setting, the best ID capacity is 10. For the panoptic setting, the best ID capacity is 10 for thing and 5 for stuff. For both R50-PAOT and SwinB-PAOT, panoptic ID strategy achieves better performance by decoupled ID banks with larger ID capacity.
% As a well-balanced point, 20 identities are proper, which retains the performance of 10 identities while runs faster and uses less memory.

\paragraph{Pyramid architecture.}
The pyramid architecture in PAOT is proposed to improve the original architecture of AOT. Here we compare two architectures and extend the LSTT in AOT-L from three to four layers (AOT-L4) for fair comparison. The results of SwinB backbone AOT and PAOT models on YouTube-VOS 2019 and VIPOSeg are in Table \ref{tab:arch}. Our pyramid architecture performs consistently better than AOT-L or AOT-L4 on different benchmarks. 
% It should be noted that these models runs almost the same FPS with the same backbone and PAOT models are even slightly faster than AOT-L and AOT-L4.

\begin{table}
\parbox{.54\linewidth}{
\centering
% \scriptsize
\fontsize{8pt}{9pt} \selectfont
\setlength{\tabcolsep}{0.5pt}
\renewcommand{\arraystretch}{0.6}
% \resizebox{\columnwidth}{!}{%
\begin{tabular}{@{}lc|c|c@{}}
\toprule
             & \multicolumn{1}{l|}{} & Y19                   & VIPOSeg             \\ \midrule
Models       & Pyramid         & $\mathcal{G}^{\star}$ & $\mathcal{G}$ \\ \midrule
SwinB-AOT-L  & \ding{55}             & 85.3                  & 73.3          \\
SwinB-AOT-L4 & \ding{55}             & 85.4                  & 74.2          \\ \midrule
\textbf{SwinB-PAOT}   & \ding{51}             & \textbf{86.5}         & \textbf{75.3} \\ \bottomrule
\end{tabular}%
% }

\caption{Comparison between AOT (no pyramid architecture) and PAOT. $\mathcal{G}^{\star}$ denotes results of all-frame test.} 
\label{tab:arch}
}
\hfill
\parbox{.45\linewidth}{
\centering
% \scriptsize
\fontsize{8pt}{9pt} \selectfont
\setlength{\tabcolsep}{0.5pt}
\renewcommand{\arraystretch}{0.8}
% \resizebox{\columnwidth}{!}{%
\begin{tabular}{@{}c|c|ccc@{}}
\toprule
          & Y19        & \multicolumn{3}{c}{VIPOSeg}                     \\ \midrule
E-LSTT    & $\mathcal{G}^{\star}$ & $\mathcal{G}$ & FPS            & Mem./GB        \\ \midrule
\ding{55} & 86.1               & \textbf{77.6}          & 6.22           & 22.00             \\
\ding{51} & \textbf{86.1}               & 77.4          & \textbf{10.45} & \textbf{11.04} \\ \bottomrule
\end{tabular}%
% }
\caption{Results before/after substituting E-LSTT for original LSTT. $\mathcal{G}^{\star}$ denotes results of all-frame test.}
\label{tab:elstt}
}
\end{table}

\paragraph{Efficient LSTT.}
E-LSTT helps the PAOT models to better balance performance and efficiency. In Table \ref{tab:elstt}, we compare the R50-PAOT models with and without E-LSTT. Models are trained with Y+D+V and evaluated on YouTube-VOS 2019 and VIPOSeg. It can be found in the table that E-LSTT causes a little performance drop, but boosts the FPS from 6 to 10, and cuts down the memory consumption from 22 GB to 11 GB.

% \usepackage{booktabs}
% \usepackage{graphicx}
% \begin{table}[]
% \small
% \resizebox{\columnwidth}{!}{%
% \begin{tabular}{@{}ccrrrr|cc@{}}
% \toprule
% IDs & $\mathcal{G}$ & \multicolumn{1}{c}{$\mathcal{G}_s$} & \multicolumn{1}{c}{$\mathcal{G}_u$} & \multicolumn{1}{l}{$\mathcal{G}^{th}$} & \multicolumn{1}{c|}{$\mathcal{G}^{sf}$} & FPS   & Mem./GB \\ \midrule
% 10          & 77.5          & 79.1                                & 75.8                                & 75.9                                   & 79.0                                    & 10.45 & 11.04       \\
% 20          & 77.4          & 78.7                                & 76.2                                & 75.7                                   & 79.1                                    & 13.32 & 8.34        \\
% 30          & 76.5          & 78.2                                & 74.8                                & 74.6                                   & 78.4                                    & 14.16 & 7.96        \\ \bottomrule
% \end{tabular}%
% }
% \caption{Evaluation results of R50-PAOT models with different identity numbers on VIPOSeg validation set.}
% \label{tab:id}
% \end{table}

\section{Conclusion}
In this paper, we explore video object segmentation in panoptic scenes and present a benchmark dataset (VIPOSeg) for it. Our VIPOSeg dataset contains exhaustive annotations, and covers a variety of real-world object categories, which are carefully divided into thing/stuff and seen/unseen subsets. Training with VIPOSeg can boost the performance of VOS methods. In addition, the benchmark is capable of evaluating VOS models comprehensively.
As a strong baseline method for panoptic VOS, 
PAOT tackles the challenges in VIPOSeg effectively by the pyramid architecture with efficient transformer and panoptic ID for panoptic object association. 
% PAOT models show superior performance with good efficiency on VIPOSeg as well as previous VOS benchmarks.
We hope our benchmark and baseline method can help the community for further research in related fields.

\section*{Acknowledgements} 
This work is supported by Major Program of National Natural Science Foundation of China (62293554) and the Fundamental Research Funds for the Central Universities (No. 226-2022-00051).
\bibliographystyle{named}
\bibliography{ijcai23}

\end{document}